# A Unified 2D Framework for DeepLesion Detection, Segmentation and Short Report Generation

Ruida Cheng[a], Tejas S. Mathai[c], Benjamin Hou[b], Qingqing Zhu[b], Zhiyong Lu[b], Matthew McAuliffe[a], Ronald M. Summers[c]

[a] Scientific Application Services, Center of Information Technology, NIH
[b] National Center for Biotechnology Information, National Library of Medicine, NIH
[c] Imaging Biomarkers and Computer-Aided Diagnosis Laboratory, Radiology and Imaging Sciences, Clinical Center, NIH

## ABSTRACT

In previous work, we integrated large language models (LLMs) into the lesion segmentation model based on the ULS23 DeepLesion dataset, using short-form findings from the reports. In this study, we developed a unified 2D lesion analysis framework that integrates LLM-based reasoning, lesion bounding box detection, segmentation, and radiology report generation from the original DeepLesion dataset. In the testing phase, we achieved relatively high lesion bounding box detection accuracy with mAP50 of 70.1%, mAP50-95 of 46.4%; Lesion segmentation performance with a Dice score of 62.6%; short report generation accuracy with BLEU_1 score of 64.3%, BLEU_4 score of 49.6%, METEOR of 34.7%, and ROUGE_L of 60.1%. In this work, we address the challenging issue of segmentation in the original DeepLesion dataset and achieve a 28.5% Dice score improvement over the nnUNet lesion segmentation model. We also integrated spatial and anatomical context into the DeepLesion short report generation. We released the implementation, dataset, and models on Github. https://github.com/ruida/2D_DeepLesion_Foundation

## 1. INTRODUCTION

DeepLesion detection, segmentation, and report generation play an essential role in medical analysis and clinical diagnosis. Recent developments in large language models (LLMs) and vision-language foundation models have created an opportunity to develop a general-purpose computer-aided diagnosis (CAD) system for medical image analysis and to enable several essential downstream clinical workflows, such as lesion detection to localize lesion regions, lesion segmentation to generate the binary lesion mask or delineate lesion boundaries for area- and volume-based measurements, and short report generation to provide a briefing on findings and impressions. However, previous studies [1, 2, 3, 4] primarily focused on lesion detection with CNN-based models and used one-hot encoding and decoding for short report generation. DeepLesion dataset-based detection and segmentation remain challenging tasks due to 1) the limited lesion annotation, 2) the spatial imbalance for tiny lesions, and 3) the scattered lesion locations in the CT image slices that are hard to identify in both detection and segmentation tasks. From the radiologist's perspective, the clinical lesion assessment, localization, delineation, and description should follow a sequential and interconnected workflow, yet most existing studies treat them as separate independent tasks. We proposed a unified 2D DeepLesion framework that integrates anatomical and semantic context into lesion detection, segmentation, and short report generation from the original DeepLesion dataset.

Lesion detection has primarily been investigated using a convolutional neural network (CNN) based architecture in the current literature. The universal lesion detector [1] introduced the DeepLesion dataset and employed a Fast R-CNN with a modified VGG-16 backbone. Later methods enhanced the lesion representation by incorporating multiscale and contextual information. In MVP-Net [5], a multi-view FPN (feature pyramid network) with different win levels was developed along with a position-aware attention network. In MULAN [2], Mask R-CNN was paired with DenseNet-121, which was truncated to address multi-task problems like object detection, segmentation, and classification. The Lesion-Harvester [3] further combined iterative hard-negative mining with a 2.5D context-enhanced CenterNet to generate 2D lesion proposals and then linked them across adjacent slices into 3D boxes. Collectively, these studies demonstrate advances in multiscale feature fusion, attention mechanisms, hard-negative mining, and 2.5D/3D contextual information. However, the Deeplesion dataset primarily focuses on small and tiny lesion regions in the 512x512 or even larger 768x768 image slices. The conventional downsampling mechanism in the encoding path can suppress and even eliminate the lesion representation on feature maps. To mitigate the issue, we intend to develop the YOLO-TLP-MOE detection model that combines spatial-preserving downsampling, multiscale attention, and sparse mixture-of-experts routines for lesion detection.

The lesion segmentation has been investigated in only a few studies. It has received much less attention than lesion detection, partly because most existing datasets provide RECIST measurements rather than annotated segmentation masks. The WSSS (Weakly-Supervised Self-Paced Segmentation) model [4] initializes image slices with RECIST line annotations on the lesion region and then propagates the prediction mask to adjacent slices in a weakly supervised manner. AHRNet [6] presented a weakly supervised lesion segmentation method by building an attention-enhanced model based on the High-Resolution Network (HRNet). MULAN [2] incorporated a mask-prediction head into multi-task Mask R-CNN to handle lesion segmentation. An active contour method [7] fused RNN feature extraction with an active contour model to segment the lesion binary mask on pathology images. In our previous work [8], we developed a text-embedded framework that encodes the DeepLesion short report findings into the Swin-UMamba decoder path to segment the lesion and achieved a relatively high segmentation performance. However, one primary constraint of this work is the lesion-centric cropped images from the ULS23 dataset. Although these methods demonstrated the feasibility of lesion delineation, many rely on ground-truth RECIST measurements, predefined lesion regions, or weakly supervised initialization. Fully automatic segmentation from the original DeepLesion image slices remains a challenging task because tiny or small lesion regions occupy only a small fraction of the overall image slice, creating a severe foreground-background imbalance. In this work, we address this issue with a detection-and-segment mechanism, where the predicted bounding boxes allocate the lesion-centric regions for segmentation. Then, the resulting binary segmentation masks are restored to the original image space.

Clinical report generation in medical imaging has been an emerging field in the current literature. Most studies formulate the task as image-to-text generation using Transformer-based or vision-language architectures, in which visual features align with or fuse with textual sequential token representations to produce a free-text report. R2Gen [9] introduced an encoder-decoder architecture with relational memory-driven Transformer for generating radiology reports. RGRG [10] proposed a region-guided report generation model that produces the final report from detected bounding-box regions. More recently, MedGemma 1.5 [11] combined the SigLIP visual encoder and the Gemma 3 language decoder for generating reports. RadVLM-GRPO [12] uses the Qwen3-VL-8B model as the visual language backbone and further refines report generation with reinforcement learning. We only briefly describe a few works in medical report generation. There is a large body of literature on LLMs for automated report prediction. Although most methods have demonstrated promising performance, a large fraction of the existing works focus on chest X-ray report generation, lacking semantic and anatomic implications from both visual grounding and meaningful text indications. These methods rely solely on whole-image-level visual features to generate the report, failing to focus on specific anatomic regions in the image. To mitigate this problem, we developed a semantic-aware report-generation model that utilizes local lesion features derived from detection and segmentation, fuses them with whole-image global visual features, and further fuses them with bounding-box location, coarse anatomical region, and rough lesion-type prompts to enhance DeepLesion short report generation.

Our contributions are as follows:

- We built a unified 2D Deeplesion framework that performs downstream tasks, such as lesion detection, segmentation, and short report generation.
- We proposed YOLO-TLP-MOE for lesion detection, which combined spatial information preservation, multiscale feature modeling, and mixture-of-experts routines. Enhanced the overall detection performance.

- We introduced a detection-guided search-and-segment mechanism that predicts bounding boxes and generates segmentation masks from cropped lesion regions to mitigate the lesion spatial imbalance challenge on the original DeepLesion image slices.
- We develop a semantic-aware report generation model that combines global and local image features and fuses them with rough anatomy region and lesion type tokens to improve overall report generation quality.

## 2. METHODS

In this section, we first give a brief overview of the proposed 2D DeepLesion framework, which consists of three main modules: YOLO-TLP-MOE for lesion detection, Swin-UMamba [13] for lesion segmentation, and R2Gen-enhanced [14] short report generation. We first describe the overall architecture, then present each module in detail.

**Overall architecture**

Given a lesion image, the proposed unified 2D framework aims to localize the lesion with a bounding box, segment the lesion, and generate a lesion-specific short report. Figure 1 illustrates the overall workflow and its main components.

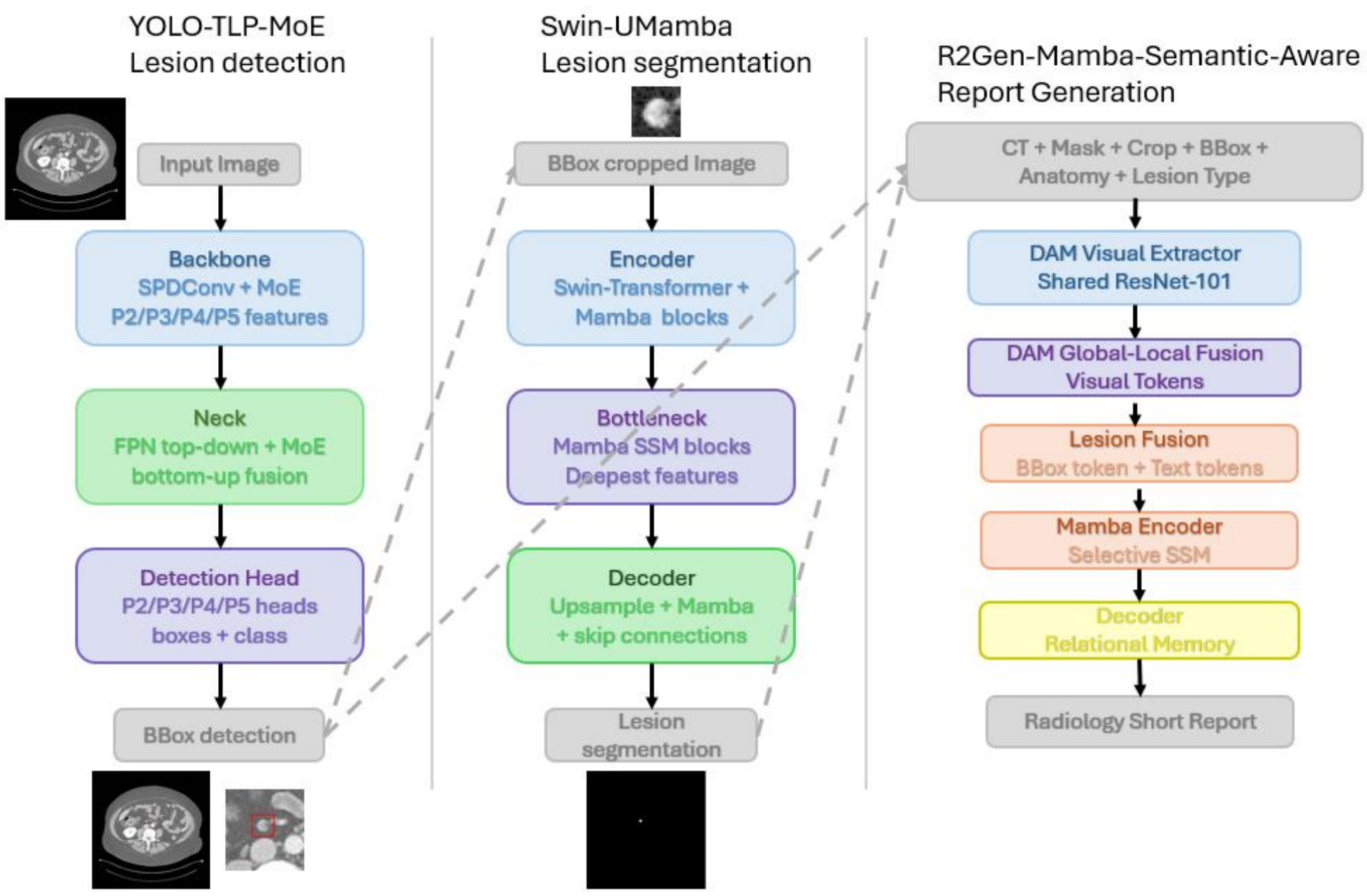

**Figure 1. Overall architecture**

**Lesion Detection**

In the DeepLesion dataset, detecting tiny lesions is challenging. Most lesions occupy a small fraction of pixels in a single 512x512 high-resolution image slice, resulting in spatial imbalance. Conventional stride convolution and pooling operations in CNNs can progressively reduce spatial resolution and consequently suppress or eliminate the lesion features during extraction in the downsampling path. For lesion detection, preserving fine-grained lesion representation with broader anatomical context appears to be a challenging task. A model trained without explicit routine or attention can easily suppress lesion activations in favor of dominant background anatomical structure. These challenges motivate us to develop a hybrid detection architecture that combines three complementary design building blocks: lossless spatial downsampling (SPDConv from YOLO-TLP [15]), adaptive feature routing (Mixture-of-Experts, MOE, from YOLO-Master [16]), and multi-scale global attention (PSA-SPPF from YOLO-TLP). We demonstrate DeepLesion detection architecture in Figure 2.

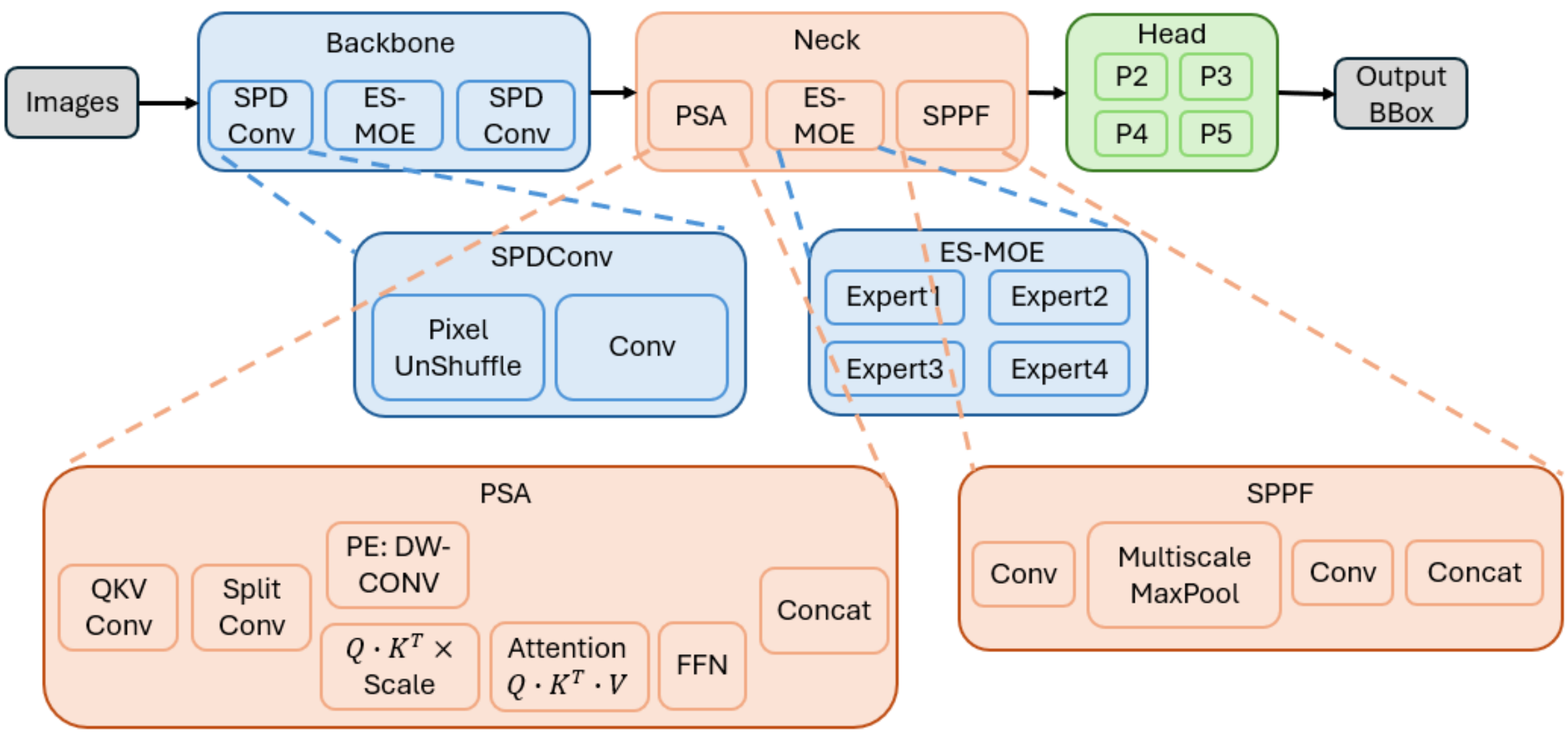

Figure 2. YOLO-TLP-MOE integrates ES-MOE blocks into the basic YOLO-TLP building blocks.

**SPDConv (Space-to-Depth Convolution):** It is a Lossless Spatial Downsampling building block. For tiny lesions, conventional stride-2 downsampling can vulnerably erase spatial information about lesion features. Preserves fine lesion details by rearranging a 2 x 2 spatial block into additional channels, mitigating lesion feature removal during downsampling.

**PSA (Position-Sensitive Attention)**: Inside the Neck building block. Uses position-sensitive attention to highlight informative tiny lesion regions and capture long-range spatial dependencies. It provides a global spatial context that complements the local precision of the SPDConv building block.

**SPPF (Spatial Pyramid Pooling):** A YOLO block within the Neck module to increase the receptive field by aggregating features at multiscale through sequential max-pooling. It can help discriminate tiny lesions from the surrounding anatomical regions.

**MoE (Mixture of Experts):** The sparse and efficient MoE module includes multiple expert subnetworks and a gating network that dynamically gates the spatial lesion regions to the most relevant experts. This MoE layer improves the representational ability.

The first MoE block inside the backbone block operates on spatially preserved features before any global context is applied, specifically determining which lesion region to extract. The second MoE block inside the Neck building block operates on contextually enriched features before multiscale pooling, focusing on how the selected lesion features are routed in the Neck. The two MoE building blocks provide specialized contextual awareness and adaptive capacity allocation for tiny lesions. The original YOLO-TLP [15] addresses the problem of preserving spatial information. The proposed architecture solves the context and multiscale aggregation problem, and the dual MoE insertion blocks address the representation capability and adaptive routing problems to enable tiny-lesion detection.

The total training loss is the weighted sum of four terms:

$$\mathcal{L} = \lambda_{box}\,\mathcal{L}_{CIoU} + \lambda_{cls}\,\mathcal{L}_{BCE} + \lambda_{dfl}\,\mathcal{L}_{DFL} + \lambda_{moe}\,\mathcal{L}_{MoE}$$

where $\mathcal{L}_{CIoU}$ is the Box loss, $\mathcal{L}_{BCE}$ is the classification loss, $\mathcal{L}_{DFL}$ is the distribution focal loss, $\mathcal{L}_{MoE}$ is the MoE land-balancing loss that computer mean routing weight per expert across the batch and spatial positions. Those $\lambda$ values are scaling weights.

**Lesion Segmentation**

Applying segmentation directly to the original 512x512 high-resolution DeepLesion dataset is challenging. Most lesion regions are small and occupy only a small fraction of the image spatially. The spatial imbalance results in a significant degradation in segmentation performance with conventional medical segmentation models. For example, nnUNet's [17] segmentation based on the original DeepLesion dataset yields only a 34% mean Dice score in the test phase, making it not an ideal solution for DeepLesion segmentation. Most likely, the tiny regions were suppressed during the downsampling step in nnUNet. Also, the lesions are scattered around the 512x512 image slice, making segmentation more challenging. To address the issue, we motivated to applying lesion detection first to generate highly likelihood lesion regions. Then we segment the cropped legion region based on the detection results. We utilize the Swin-UMamba [13] to segment the cropped lesion regions. Based on the original DeepLesion dataset bounding boxes and the grab-cut-generated lesion labels, we crop the lesion region for both images and labels. Those cropped image-label pairs vary in image size.

After curating the crop lesion region dataset, we train using the Swin-UMamba architecture. We chose Swin-UMamba as the segmentation backbone due to its outstanding performance in the latest medical image segmentation task, surpassing traditional Convolutional Neural Networks (CNNs), Vision Transformers (ViTs) [18], and even other Mamba-based models, such as U-Mamba and its derivatives. Architecture-wise, the Swin-UMamba has a few advantages over nnUNet: 1) ability to handle long-range dependencies more effectively; 2) integration of UMamba state space attention mechanism, which enables more efficient and expressive sequential modeling. Thus, Swin-UMamba is better suited for segmentation than nnUNet. After generating the prediction labels for the cropped lesion regions, we merge or patch the predicted labels back into the original 512x512 label images for evaluation.

**Short Report Generation**

Based on the detected lesion bounding box, we integrate the R2Gen-Mamba [14] architecture with the Describe Anything Model (DAM) [19] and a Bounding Box (BBox) anatomy-aware, lesion-type-aware architecture to generate the lesion short report. The original R2Gen-Mamba [14] generates the medical report from the entire 2D image and relies fully on visual features. This design ignores the related structured clinical metadata. In the DeepLesion dataset, each lesion can be roughly categorized into one of the 11 anatomical regions (Table 1) and one of the 10 lesion types (Table 2), each contributing specific semantic cues to short report generation.

We propose the R2Gen-Mamba-Semantic-Aware framework (Figure 3), which jointly models lesion-focused visual representation, anatomical location, and lesion type to generate the short report. The full-scale CT image and lesion mask, together with the lesion region image (BBox-cropped) and mask, are processed by a shared ResNet-101 visual extractor to obtain both global and local visual features. Global and local visual features are integrated through gated global-to-local fusion (attention) to produce lesion-aware DAM visual tokens. In parallel, the normalized bounding box coordinates are encoded as a spatial feature token, while the combined anatomy and lesion type text are encoded as semantic tokens. LesionFusion combines the DAM visual tokens, bounding-box token, and semantic text tokens into a conditioned visual sequence. The sequence is then processed by the Mamba selective state-space encoder and the relation-memory decoder to generate the final short report. Thus, the anatomy and lesion types provide the rough semantic conditioning for short report generation.

After detecting the bounding box from the YOLO-TLP-MOE model, we aligned the BBox with the pre-segmented anatomical region label to identify the rough anatomical region. During inference time, the BBox, rough anatomical region, and lesion type have been encoded as spatial and semantic context, which is hard to recover from images alone (i.e., the original R2Gen-Mamba [14]) or, particularly, for small or ambiguous lesions. The anatomical region and lesion type provide the encoder with more detailed semantic tokens, which can be considered soft conditioning input. The decoder is adapted to the semantic-conditioned features. The report generation step may be more biased toward that semantic context. So, the anatomy and lesion-type terms do not enter the decoder as labels, but they can guide the decoder through the encoded representation. This mechanism ensures that no data leaks to the decoder (Relational Memory). We argue that grounding short report generation in explicit lesion location (BBox) and semantic context (rough anatomical region and lesion type) could improve the quality of predicted reports.

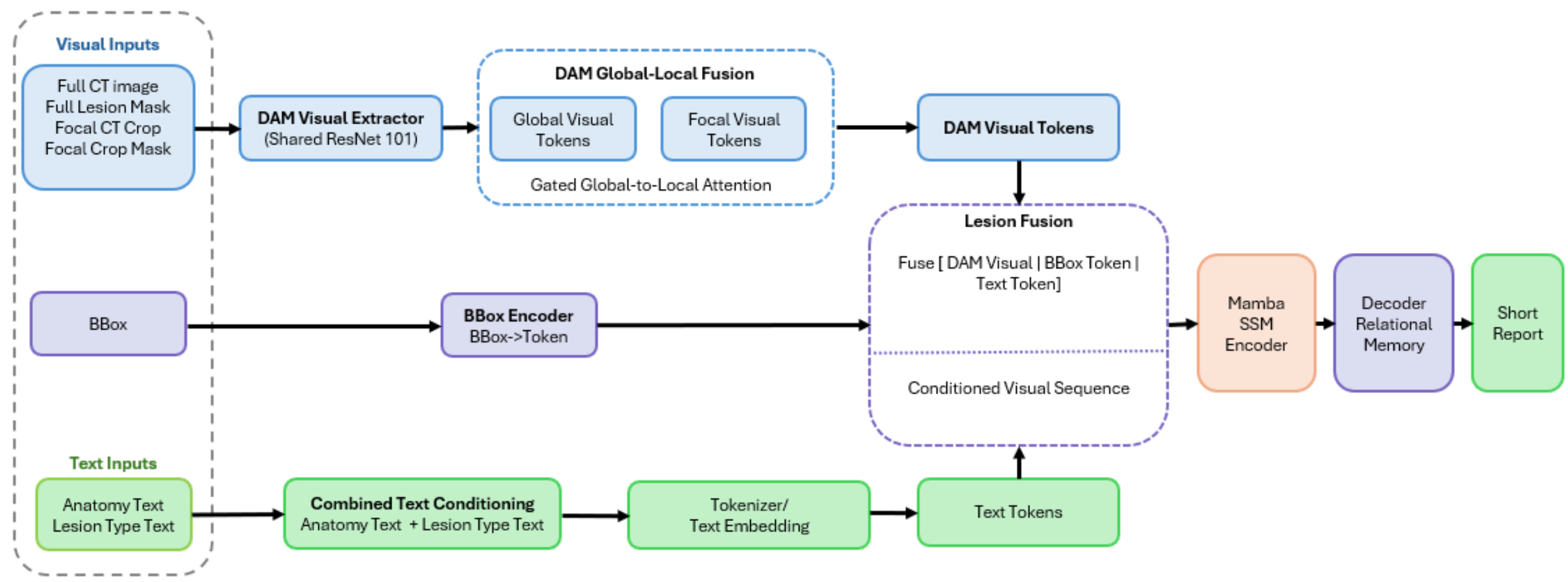

Figure 3. R2Gen-Mamba-Semantic-Aware framework: DAM + Anatomy + Lesion Type guided short report generation

To align the detection bounding box with the anatomical region, we used the TotalSegmentator [20] to generate 3D segmentation maps of rough anatomical structures, then extracted the corresponding 2D slice from each 3D segmentation to match the DeepLesion 2D slice representation. Then, we derive an algorithm to allocate the detected bounding box coordinates to approximate organ regions, e.g., lung or liver. Figure 4 demonstrates the 2D image slice with detected bounding boxes and the corresponding segmentation map. The pseudo-code (Algorithm 1 in Appendix) explains the algorithm to aligning the bounding box with a rough anatomical region, then mapping it to the 11 anatomy regions.

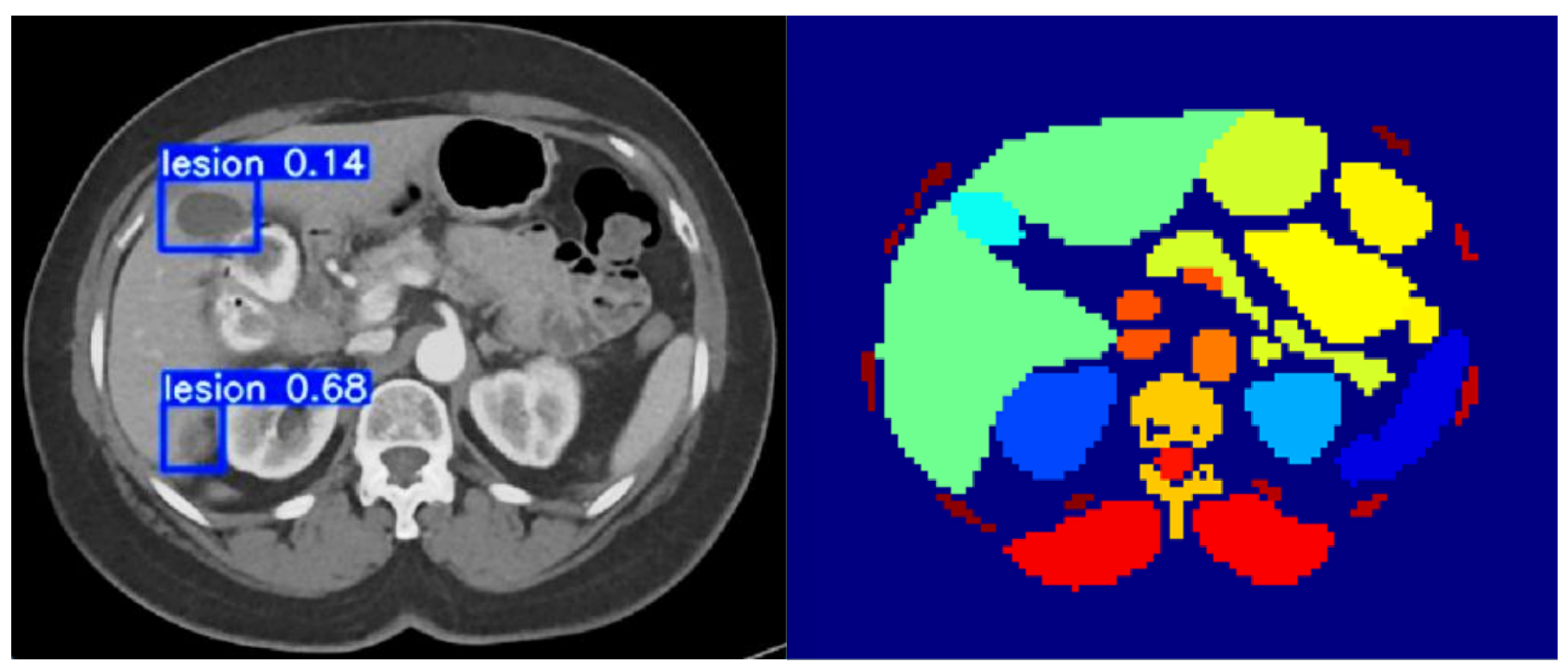

Figure 4. Align bounding box with the rough anatomical region ID

**Table 1. DeepLesion Rough Anatomy Regions**

| ID | 1 | 2 | 3 | 4 | 5 | 6 |
| --- | --- | --- | --- | --- | --- | --- |
| **Name** | lung | liver | kidney | adrenal | abdomen | pelvis |
| **ID** | 7 | 8 | 9 | 10 | 11 | |
| **Name** | chest | brain_and_neck | spine | bone | unknown | |

**Table 2. DeepLesion Rough Lesion Types**

| ID | 1 | 2 | 3 | 4 | 5 |
| --- | --- | --- | --- | --- | --- |
| **Name** | lesion | lymph-node | soft_tissue | nodule | mass |
| **ID** | 6 | 7 | 8 | 9 | 10 |
| **Name** | opacity | cystic | enhancing | calcified | metastatic |

During inference, the lesion bounding box predicted by YOLO-TLP-MOE is used to define the focal lesion crop and localize the lesion within the anatomical segmentation map. The detected bounding box is aligned with the corresponding 2D TotalSegmentator label map, and the anatomical structure with the greatest overlap inside the box is mapped to one of the predefined rough anatomical regions. The lesion type is predicted from the focal lesion region using a lesion-type classification module. The resulting anatomy description and lesion type are converted into short-text phrases, such as “lung” and “nodule,” and combined into a single conditioning sequence. The text sequence, joined with DAM visual tokens, is used to generate the final short report.

## 3. EXPERIMENTS

From the original DeepLesion dataset, we extract the 2D PNG image slice from the 3D images. We use GrabCut [21] to generate DeepLesion binary masks from the corresponding bounding boxes on the 2D image slice. Additionally, we generate two modality image slices by using different Win-Level thresholds for each 2D image slice. We split the train, val, and test datasets by patient ID and ensure there are no overlapping image-mask slice pairs across them. Table 3 illustrates the quantitative number of the train, val, and test datasets for YOLO-TLP-MOE DeepLesion detection. Table 4 demonstrates the train, val, and test distribution for Swin-UMamba cropped DeepLesion segmentation, which crops the bounding box around the lesion region for both image and mask, and trains the Swin-UMamba only on the cropped image mask pairs. Table 5 shows the train, val, and test splits for R2Gen-Mamba-Semantic-Aware short report generation. The numbers in Table 5 are constrained by the short report entries of the original DeepLesion dataset. Figure 5 illustrates two samples of the paired image, mask and short report.

Table 3. DeepLesion dataset training, validation, test case numbers for YOLO-TLP-MOE DeepLesion detection.

| | Training | Validation | Test |
| --- | --- | --- | --- |
| **DeepLesion detection dataset** | 51207 | 6392 | 6399 |

Table 4. DeepLesion dataset training, validation, test case numbers for Swin-UMamba Cropped DeepLesion segmentation.

| | Training | Validation | Test |
|---|---|---|---|
| **DeepLesion segmentation dataset** | 18233 | 4558 | 4588 |

Table 5. DeepLesion dataset training, validation, test case numbers for R2Gen-Mamba-Semantic aware short report generation.

| | Training | Validation | Test |
|---|---|---|---|
| **DeepLesion segmentation dataset** | 16667 | 2058 | 2026 |

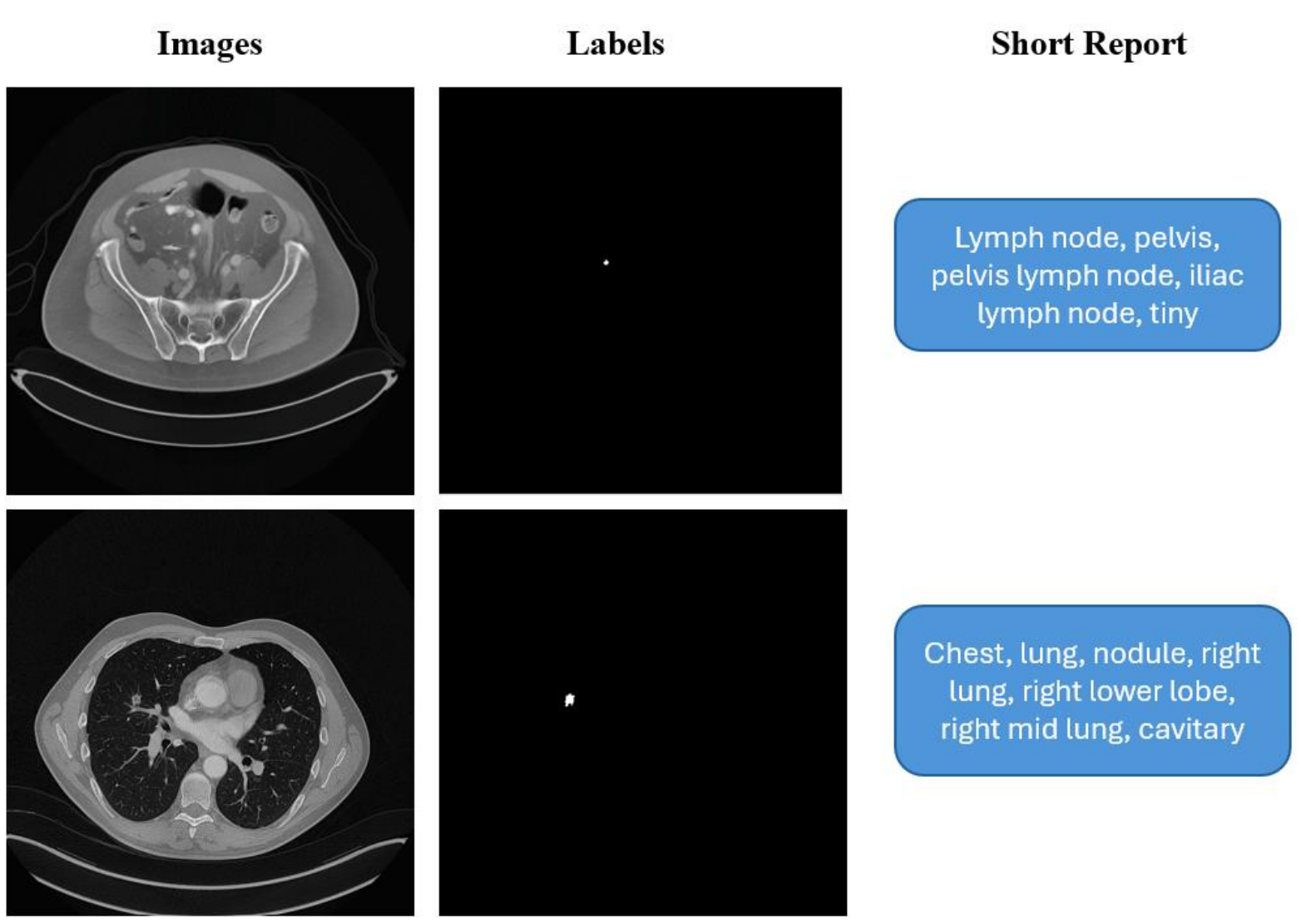

Figure 5. Paired image, label, and corresponding short form report.

For short report generation, we proposed the R2Gen-Mamba-Anatomy-Aware model, which utilizes ground-truth lesion masks, bounding boxes, ground-truth anatomy tokens, and lesion-type tokens extracted from the short reports as conditioning inputs, while the corresponding ground-truth short radiology report serves as the report target. During the testing phase, the YOLO-TLP-MOE predicted bounding boxes, the Swin-UMamba predicted lesion masks, the TotalSegmentator predicted anatomy tokens, and the Dual-head classifier predicted lesion type tokens as the conditioning inputs to generate the final short reports, and were evaluated with the corresponding ground-truth short reports. We leave the detailed explanation of the training and testing scenario for the short report generation in the Appendix section.

We compare the proposed 2D unified DeepLesion framework component-wise with the latest models. For the detection, we compared the YOLO-TLP-MOE with YOLO-TLP and YOLO-MOE. For segmentation, we compared full-resolution results from nnUNet[17], Swin-UMamba [13], and KEN [22]. For short report generation, we compared the R2Gen-Mamba-Semantric-aware model with Qwen3-VL [23] and MedGemma 1.5 [11]. All models are trained on a single NVIDIA A100 GPU with 80 GB of memory. During the testing phase, we use general metrics, such as the mean Dice coefficient for segmentation, mAP50 for detection, and BLEU scores for short report generation.

# 4. RESULTS

For the lesion Bounding Box (BBox) Detection module, we compared the proposed YOLO-TLP-MOE models with three baseline models, YOLO-TLP [15] and YOLO-Master (MOE) [16], and YOLO-12 [24]. We utilize the standard YOLO detection metrics: mAP50, mAP50:95, Precision, and Recall comparing the detection performance (Table 6). Figure 6 demonstrates a few comparison cases for lesion detection.

Table 6. Comparison of bounding box detection performance for DeepLesion dataset

| | Precision | Recall | mAP50 | mAP50:95 |
|---|---|---|---|---|
| YOLO-12 [24] | 0.570 | 0.514 | 0.505 | 0.291 |
| YOLO-TLP [15] | 0.659 | 0.640 | 0.673 | 0.442 |
| YOLO-Master (MOE) [16] | 0.660 | 0.643 | 0.672 | 0.422 |
| **YOLO-TLP-MOE** | **0.665** | **0.674** | **0.701** | **0.464** |

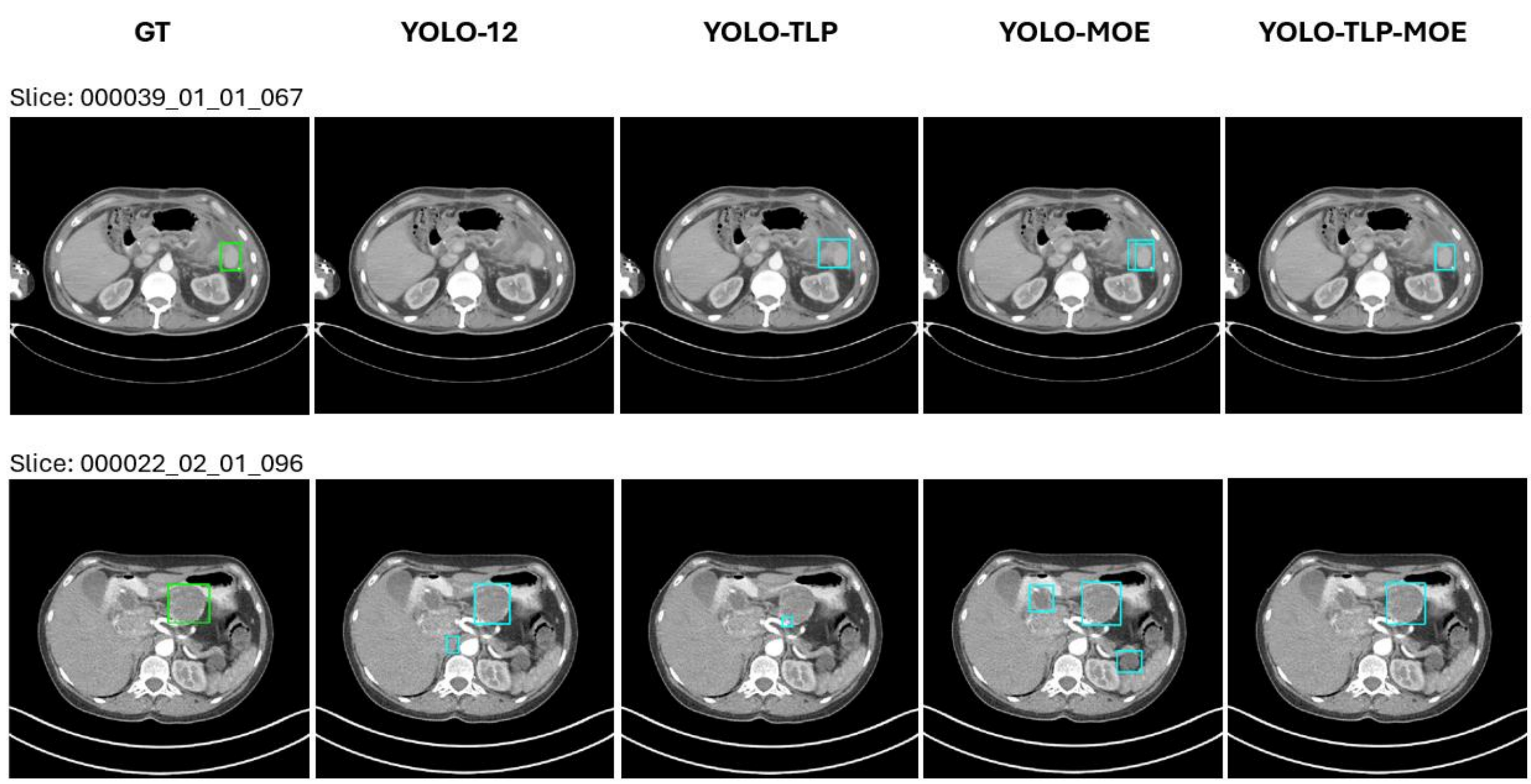

Figure 6. Comparison of predicted lesion bounding box detection

Based on the YOLO-TLP-MOE detected bounding boxes during the testing phase, we test the Swin-UMamba segmentation model on the cropped lesion regions and merge or patch the cropped segmentation map back into the original 512x512 image space. We compare segmentation performance (Table 7) with nnUNet, Swin-UMamba, and KEN using the mean Dice score—all the three models apply direct segmentation to the original DeepLesion dataset. Figure 7 shows the visual comparison of DeepLesion segmentation results.

Table 7. Comparison of lesion segmentation performance for DeepLesion dataset

| | Mean Dice | Hausdorff95 | Mean IoU |
|---|---|---|---|
| nnUNet [17] | 0.341±0.413 | 55.711±70.438 | 0.298±0.377 |
| KEN [22] | 0.529±0.002 | 224.144±1.752 | 0.507±0.001 |
| Swin-UMamba [13] | 0.401±0.426 | 57.244±74.119 | 0.356±0.397 |
| **Swin-UMamba crop & merging** | **0.626±0.416** | **39.036±65.973** | **0.576±0.397** |

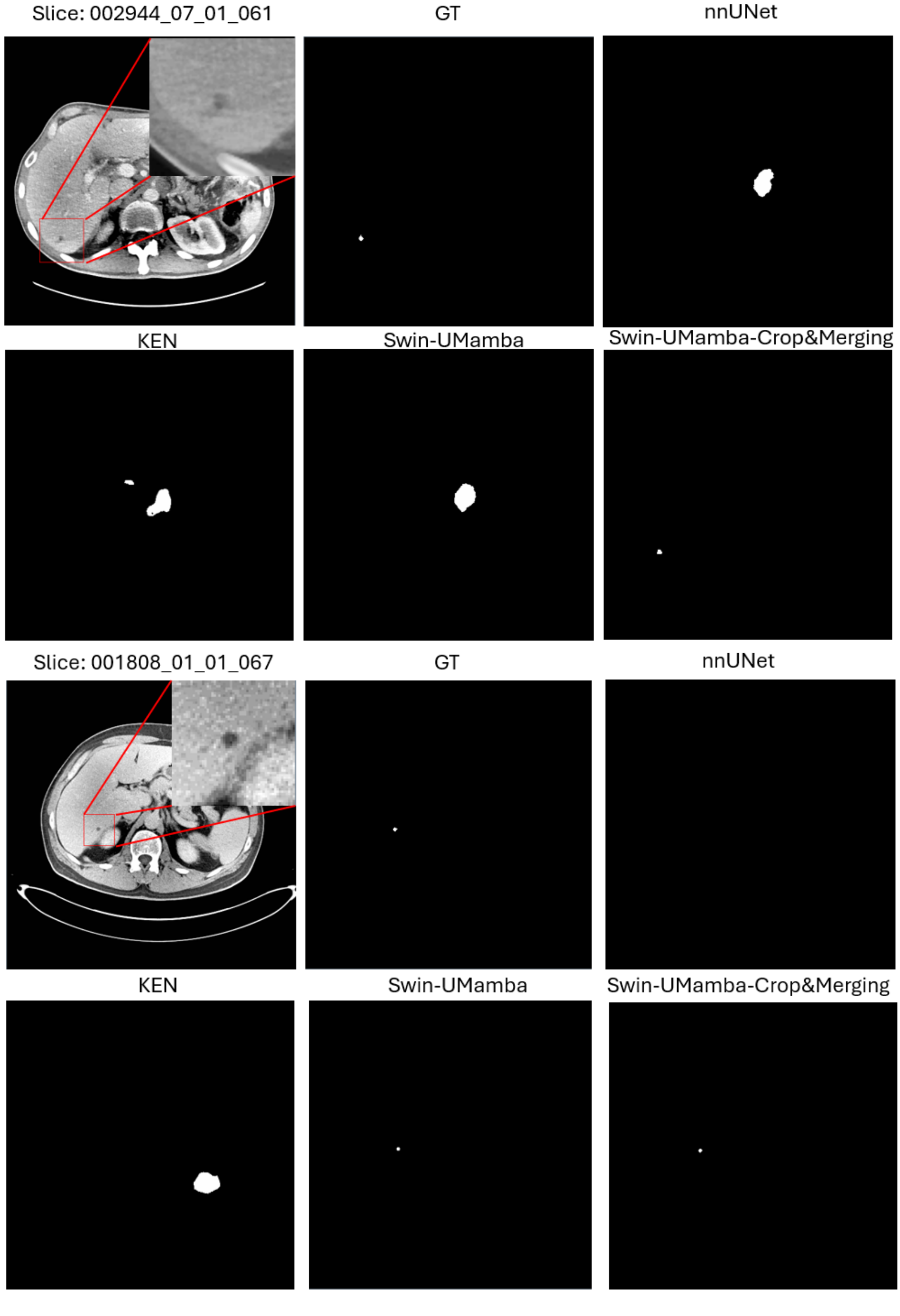

Slice: 002944_07_01_061
GT
nnUNet
KEN
Swin-UMamba
Swin-UMamba-Crop&Merging
Slice: 001808_01_01_067
GT
nnUNet
KEN
Swin-UMamba
Swin-UMamba-Crop&Merging

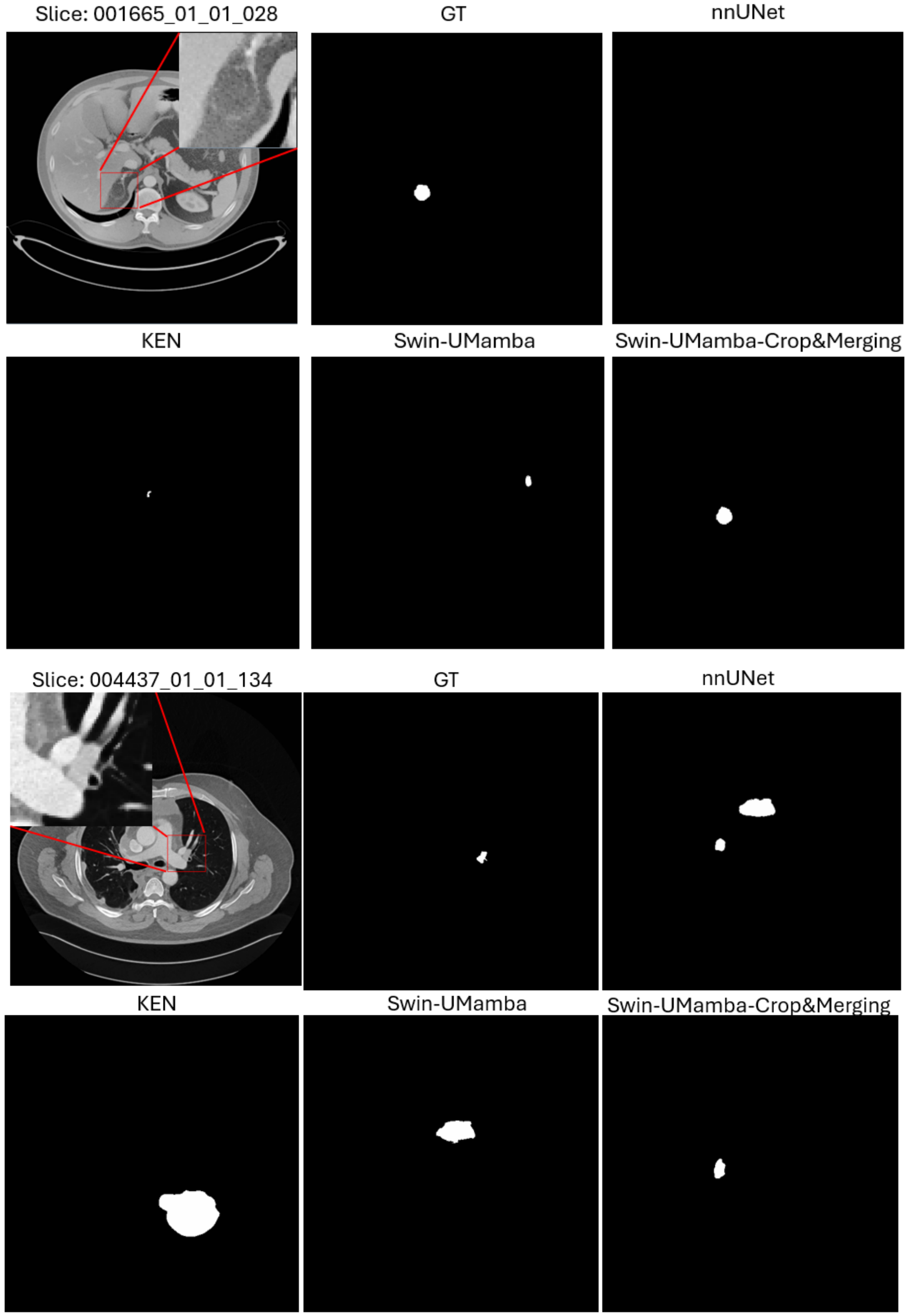
Slice: 001665_01_01_028
GT
nnUNet
KEN
Swin-UMamba
Swin-UMamba-Crop&Merging
Slice: 004437_01_01_134
GT
nnUNet
KEN
Swin-UMamba
Swin-UMamba-Crop&Merging

Figure 7. Qualitative comparison of the predicted lesion segmentation results. The ground truth segmentation is followed by each model's segmentation result (Table 7). The GT represents the ground truth. Slice 004437_01_01_134 represents the missing segmentation case.

The R2Gen-Mamba-Anatomy-Aware model adds anatomy term, lesion type and the detected bounding box tokens to the R2Gen-Mamba architecture's encoder to improve DeepLesion short report generation. To evaluate the performance of short report generation, we compare our proposed model with the latest state-of-the-art medical report generation models, MedGemma 1.5 and Qwen3-VL. As shown in Table 8, our proposed R2Gen-Mamba-Anatomy-Aware model outperforms MedGemma and Qwen3-VL by a large margin. We fine-tuned the MedGemma and Qwen3-VL models using the same train, validation, and test datasets of the DeepLesion. We implemented the Anatomy-Aware mechanism into MedGemma 1.5 and Qwen3-VL encoders and conduct the ablation experiments with Anatomy-Aware MedGemma 1.5 and Qwen3-VL. Figure 8 compares a few cases for the short report generation in testing phase.

Table 8. Comparison of short report generation performance for DeepLesion dataset

| | BLEU_1 | BLEU_4 | METEOR | ROUGE_L | Precision | Recall | F1 |
|---|---|---|---|---|---|---|---|
| MedGemma 1.5 [11] | 0.334 | 0.181 | 0.219 | 0.367 | 0.351 | 0.441 | 0.370 |
| Qwen3-VL [23] | 0.344 | 0.156 | 0.353 | 0.398 | 0.435 | 0.451 | 0.413 |
| R2Gen-Mamba [14] | 0.563 | 0.307 | 0.251 | 0.545 | 0.651 | 0.602 | 0.595 |
| MedGemma1.5-Anatomy-Aware | 0.447 | 0.297 | 0.344 | 0.472 | 0.497 | 0.607 | 0.520 |
| Qwen3-VL-Anatomy-Aware | 0.391 | 0.170 | **0.407** | 0.443 | 0.469 | 0.526 | 0.465 |
| **R2Gen-Mamba-Anatomy-Aware** | **0.643** | **0.496** | 0.347 | **0.601** | **0.704** | **0.666** | **0.650** |

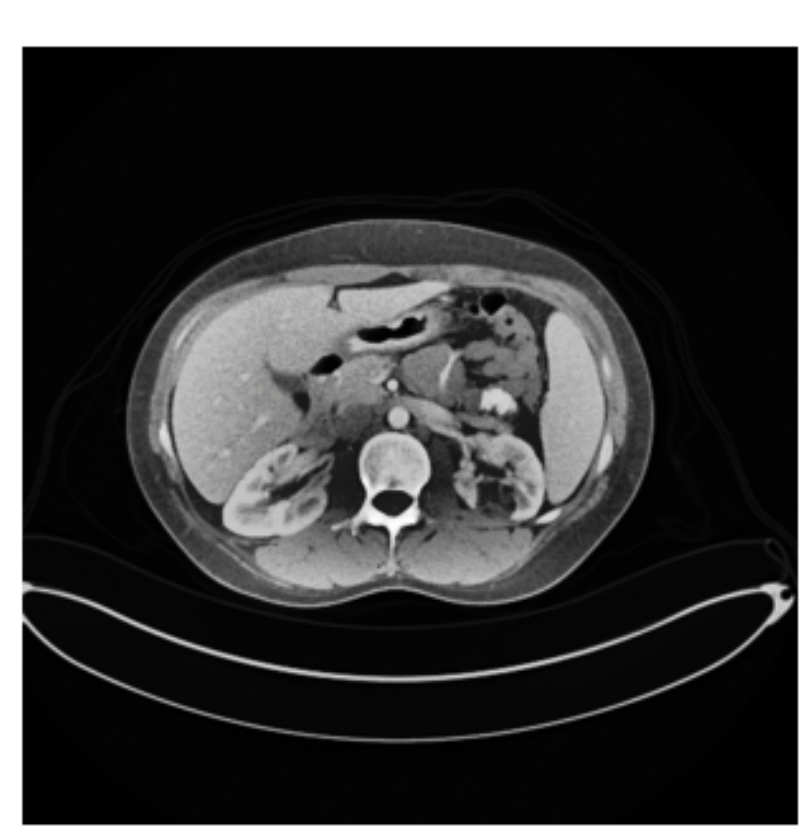

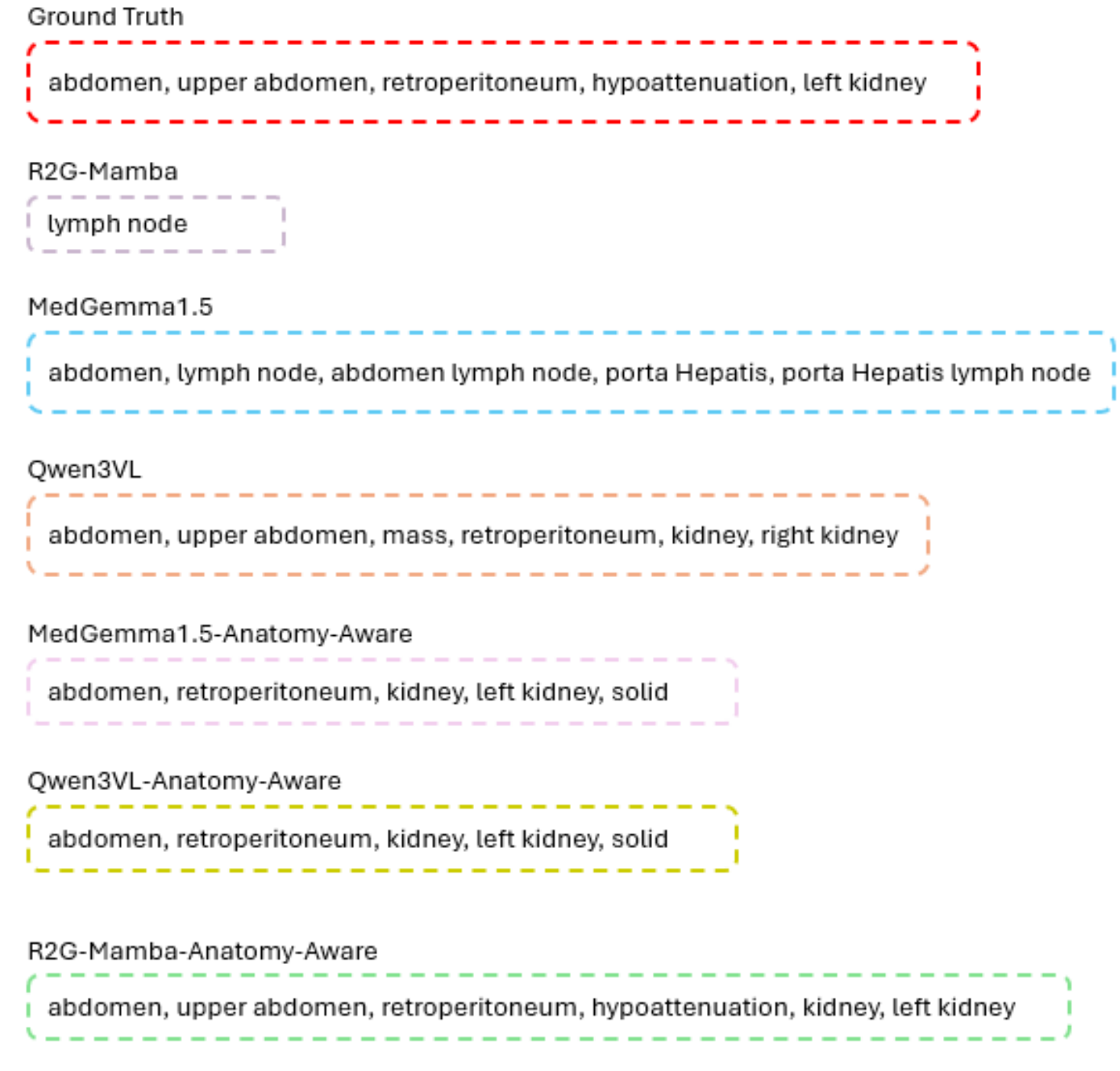

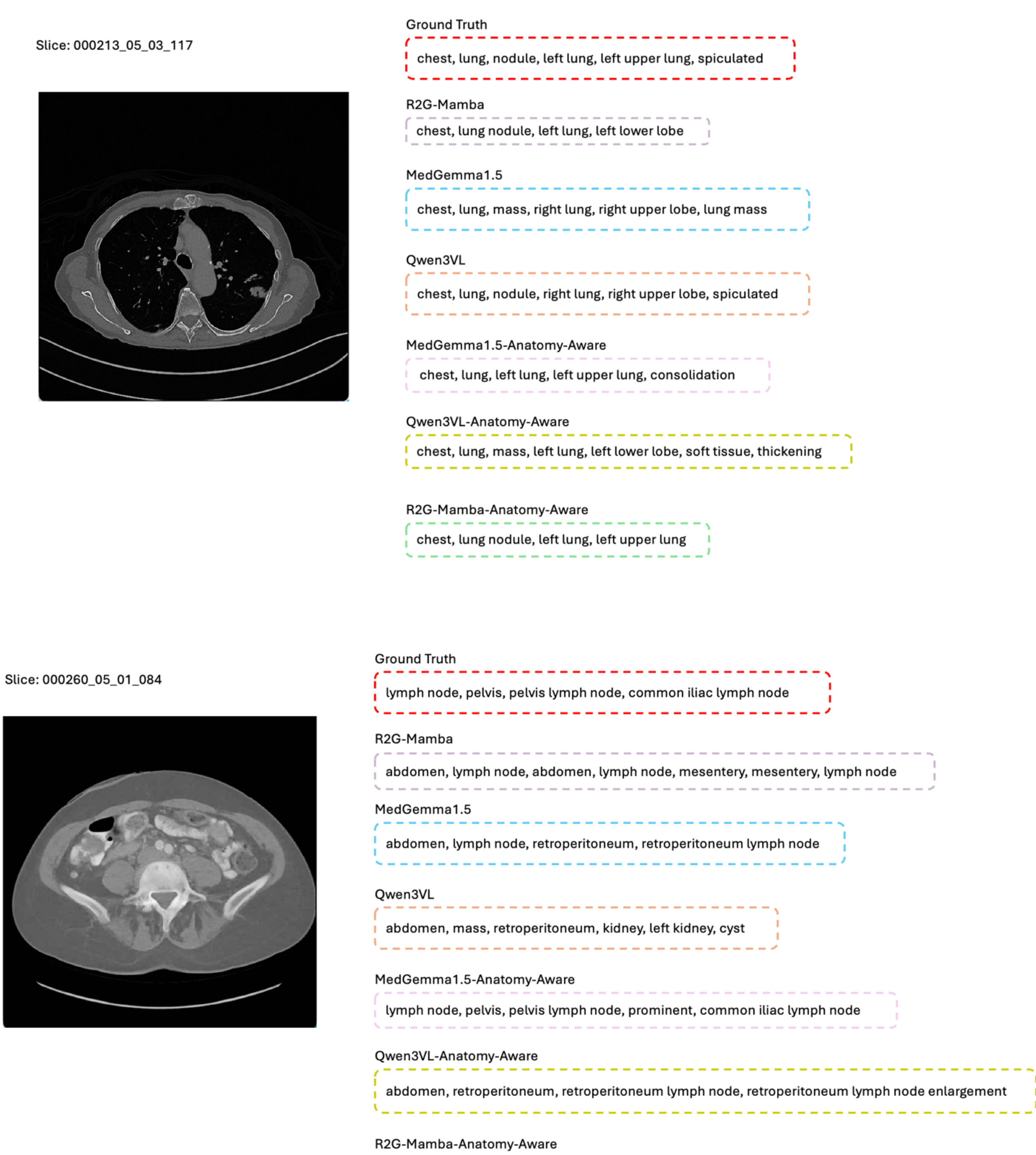

Figure 8. Comparison of the short report generation

# 5. DISCUSSION

In this study, we proposed a simplified version of the 2D foundation model -- a unified 2D framework that performs tiny lesion detection, segmentation, and report generation using the original DeepLesion dataset. The experimental results demonstrate substantial performance gains in downstream DeepLesion analysis tasks. YOLO-TLP-MOE yields the best lesion detection performance over the other models, with mAP50 of 70.1% and mAP50:95 of 46.4%, which indicates that spatial-preserving downsampling and mixture-of-experts building blocks can benefit tiny lesion detection routines. More importantly, the detection-guided search-and-segment mechanism (Swin-UMamba cropped and merged) outperforms the conventional direct segmentation models with a mean Dice score of 62.6%, compared with 34.1% for nnUNet and 40.1% for Swin-UMamba for the original 512 × 512 DeepLesion image slices. This finding indicates that focusing the segmentation model on predicted likelihood lesion regions can mitigate the severe tiny-lesion foreground-background imbalance issue in the original DeepLesion image space. This overall detection and segmentation strategy is fully automatic without any initial human prompt or annotation. For the report generation, we proposed the anatomy-aware R2Gen-Mamba model that combines global and local lesion visual feature representations with bounding-box, coarse anatomical region, and lesion type embedding, which achieved the best overall DeepLesion short report generation performance, with BLEU-1 of 0.643, BLEU-4 of 0.496, ROUGE-L of 0.601, and F1 of 0.650. We obtain a substantial performance margin improvement over Qwen3-VL and Gemma 1.5 models with 0.299 and 0.309, respectively. We also find an interesting trend: with an anatomy-aware integration into the baseline Qwen3-VL, Gemma 1.5, and R2Gen-Mamba models, report-generation performance improves by 0.05, 0.11, and 0.05 in BLEU-1 scores, respectively. In other words, this anatomy-aware mechanism explores the feasibility of using rough anatomy lesion type and lesion focal features as prompts to improve report-generation quality. However, this study has certain limitations. The segmentation masks are generated using GrabCut rather than expert delineation; the segmentation and report generation are highly dependent on the 2D lesion bounding box detection. In case of missing detection (i.e., Figure 7, slice 004437_01_01_134), the misprediction error can propagate into subsequent downstream tasks and limit the accuracy of semantic conditioning. Future work should focus on evaluating the DeepLesion dataset from 3D images, incorporating expert annotation on lesion data, integrating a 3D volumetric lesion-searching mechanism, improving 3D lesion type and anatomical localization, and developing unified DeepLesion foundation model in 3D.

# 6. CONCLUSION

We proposed an integrated DeepLesion 2D framework for tiny-lesion detection, detection-assisted segmentation, and lesion focus short-report synthesis from the original-resolution DeepLesion 2D CT image slices. The proposed YOLO-TLP-MOE detector was capable of retaining fine spatial detail information in layer-wise feature propagation along with multiscale and mixture-of-experts routing. The search-and-segment strategy depended on the predicted lesion bounding boxes to work around the tiny-lesion spatial imbalance issue and advance lesion delineation. The report-generation module was motivated by an anatomy- and semantically aware mechanism to combine global image features, localized lesion features, bounding-box information, coarse anatomical region, and lesion type to generate semantically grounded short reports. The proposed framework achieved an mAP50 of 0.701 for lesion detection, a mean Dice score of 62.6% for lesion segmentation, and BLEU-1 and BLEU-4 scores of 0.643 and 0.496 for report generation. These results support the feasibility of effectively downstreaming lesion localization, delineation, and reporting tasks within a unified framework and suggest that spatially focused, semantically conditioned modeling can improve automated lesion report generation.

# 7. APPENDIX

**Algorithm of the Bounding Box Alignment with TotalSegmentator-Generated Segmentation Mask**

**Algorithm 1:** Align Bounding Boxes with Rough Anatomical IDs

**Input:** predicted-label directory $P$, mask PNG directory $M$, expansion ratio $r$, TotalSegmentator class map $L$.
**Output:** Results $R$ = {(image_id, box_index, bbox, rough_id, rough_name)}.

```
1  Initialize R as an empty list
2  Load the TotalSegmentator class map L ← LoadTotalSegClassMap()
3  ▷ Match Predicted Bounding Boxes with Segmentation Masks
4  foreach txt_path in P do
5  |   image_id ← GetImageID(txt_path)
6  |   mask_path ← FindCorrespondingMask(M, image_id)
7  |   Mask ← ReadPNG(mask_path)  // shape [H, W]
8  |   boxes ← ParseYOLO(txt_path)
9  |   foreach box_index, box in boxes do
10 |   |   x1, y1, x2, y2 ← YOLOToPixel(box, H, W, r)
11 |   |   crop ← Mask[y1:y2, x1:x2]
12 |   |   counts ← Counter(crop)  // {label_id → pixel_count}
13 |   |   rough_votes ← {}
14 |   |   ▷ Vote for Rough Anatomical ID from Overlapping Mask Labels
15 |   |   foreach label_id, n in counts do
16 |   |   |   anatomy_name ← L[label_id]
17 |   |   |   rough_id ← KeywordMatch(anatomy_name)  // rough_id ∈ {0, 1, ..., 12}
18 |   |   |   rough_votes[rough_id] ← rough_votes[rough_id] + n
19 |   |   rough_id ← ArgMax(rough_votes)
20 |   |   rough_name ← RoughIDToName(rough_id)
21 |   |   bbox ← {x1, y1, x2, y2}
22 |   |   Append {image_id, box_index, bbox, rough_id, rough_name} to R
23 return R
```

The bounding box alignment to the anatomical region algorithm's core idea is simple. For each YOLO-TLP-MOE detected bounding box, crop the TotalSegmentator-generated segmentation mask map into the box, and count the pixels that belong to each anatomical label. Finally, map those labels to the 11 rough anatomical classes via keyword matching and assign the region with the most pixels.

### Short Report Generation Training and Testing Scenarios

Training Process

During the training phase, each DeepLesion sample from the DeepLesion dataset consists of a CT image, a ground-truth lesion segmentation mask, a bounding box-derived cropped lesion image and corresponding mask, a ground-truth anatomy annotation (extracted from the ground truth short report), and a ground-truth lesion type label (extracted from the ground truth short report). The DeepLesion segmentation mask and bounding box provide more accurate lesion anatomy localization, while the anatomy and lesion type together provide complementary, semantically related information.

The original DeepLesion CT image and ground-truth lesion masks were used to construct the global- and lesion-focused visual inputs. The bounding box from the original DeepLesion defined the local lesion region and was used to extract an enlarged lesion-centric crop. The global image, global mask, focal image crop, and focal mask crop were then processed by the DAM visual pathway to obtain lesion and anatomy-aware visual representations. Simultaneously, the ground-truth anatomy was converted into anatomy tokens, and the ground-truth lesion category was encoded as a lesion type token. Those semantic tokens then fused with the lesion-aware visual features and bounding-box representation. The combined multimodal representation guides the R2Gen-Mamba decoder in generating the short lesion report.

The generated token sequence was compared with the ground-truth short report using a language generation objective function. The resulting loss was back-propagated through the report decoder, semantic modules, DAM fusion components, and trainable visual layers. Thus, the model learned to more accurately associate localized lesion appearance with ground-truth semantic information in the corresponding radiology report. Practically, the ground-truth report served as the prediction target, whereas the ground-truth mask, bounding box, anatomy tokens, and lesion type tokens served as conditioning inputs during training.

Lesion Type Classification

In the training phase, a dual-head image classifier with a ResNet backbone performs rough lesion-type and attribute prediction. For each training instance, the ground-truth bounding-box-based cropped image and lesion mask were extracted, resized to the general image classification size of 224 x 224, and the cropped image was normalized. The ResNet visual backbone generated a feature representation that was passed to two classification heads: one to predict the rough lesion type, and the other to predict a complementary rough lesion attribute. The classifier optimization utilized class-weighted cross-entropy losses for both heads, with the attribute loss weighted relative to the primary classification loss to alleviate class imbalance.

During the testing phase, the same preprocessing and classification procedure was applied without data augmentation. For realistic inference, lesion crops were generated from the YOLO-TLP-MOE predicted bounding boxes. The trained classifier produced rough lesion-type and lesion-attribute predictions, which were converted into lesion-type conditioning tokens for the report-generation model. On the validation set, the classifier achieved 60.68% lesion type accuracy, 81.77% lesion attribute accuracy, and 50.72% joint accuracy. On the testing set, the corresponding accuracies were 56.56%, 80.09%, and 47.87%, respectively.

Testing Process

During testing, the proposed R2Gen-Mamba-Anatomy-Aware received YOLO-TLP-MOE predicted bounding boxes, Swin-UMamba predicted lesion masks, TotalSegmentator derived predicted anatomy region tokens, and dual-head classifier predicted rough lesion type and attribute tokens as conditioning inputs. The model generated short report without access to ground-truth localization or semantic labels, and the generated reports were evaluation against the corresponding ground-truth short reports.

## ACKNOWLEDGEMENTS

This research was supported [in part] by the Intramural Research Program of the National Institutes of Health (NIH). The contributions of the NIH author(s) are considered Works of the United States Government. The findings and conclusions presented in this paper are those of the author(s) and do not necessarily reflect the views of the NIH or the U.S. Department of Health and Human Services.